\documentclass[sigconf]{aamas} 

\usepackage{balance} % for balancing columns on the final page
\usepackage{colortbl}
\usepackage{tikz} 
\usepackage{algorithm}
\usepackage{algorithmic}
\usepackage{cuted}

\usetikzlibrary{arrows,decorations.pathmorphing,backgrounds,fit,positioning,shapes.symbols,chains}
\tikzset{box1/.style={draw=black, thick, rectangle, minimum height=1.4cm, minimum width=3cm }}

\setcopyright{iw3c2w3}
\acmConference[]{MILES Working paper - to appear in AAMAS ALA 2021 workshop}{LAMSADE}{Dauphine University}{Machine Learning Group}
\copyrightyear{2021}
\acmYear{2021}
\acmDOI{}
\acmPrice{}
\acmISBN{}

\acmSubmissionID{56}

\title[AAMAS-2021 Formatting Instructions]{Adaptive learning for financial markets \\
mixing model-based and model-free RL for volatility targeting
}

\author{Eric Benhamou}
\affiliation{  \institution{AI for Alpha}}
\email{eric.benhamou@aiforalpha.com}

\author{David Saltiel}
\affiliation{\institution{AI for Alpha}}
\email{david.saltiel@aiforalpha.com}

\author{Serge Tabachnik}
\affiliation{\institution{Lombard Odier}}
\email{s.tabachnik@lombardodier.com}

\author{Sui Kai Wong}
\affiliation{\institution{Lombard Odier}}
\email{sk.wong@lombardodier.com}

\author{François Chareyron}
\affiliation{\institution{Lombard Odier}}
\email{f.chareyron@lombardodier.com}

\begin{abstract}
 Model-Free Reinforcement Learning has achieved meaningful results in stable environments 
but, to this day, it remains problematic in regime changing environments like financial 
markets. In contrast, model-based RL is able to capture some fundamental and dynamical 
concepts of the environment but suffer from cognitive bias. In this work, we propose 
to combine the best of the two techniques by selecting various model-based approaches 
thanks to Model-Free Deep Reinforcement Learning. Using not only past performance 
and volatility, we include additional contextual information such as macro and risk 
appetite signals to account for implicit regime changes. We also adapt traditional 
RL methods to real-life situations by considering only past data for the training 
sets. Hence, we cannot use future information in our training data set as implied 
by K-fold cross validation. Building on traditional statistical methods, we use the 
traditional "walk-forward analysis", which is defined by successive training and 
testing based on expanding periods, to assert the robustness of the resulting agent. 

Finally, we present the concept of statistical difference's significance based on 
a two-tailed T-test, to highlight the ways in which our models differ from more traditional 
ones. Our experimental results show that our approach outperforms traditional financial 
baseline portfolio models such as the Markowitz model in almost all evaluation metrics 
commonly used in financial mathematics, namely net performance, Sharpe and Sortino 
ratios, maximum drawdown, maximum drawdown over volatility.
\end{abstract}


\newcommand{\BibTeX}{\rm B\kern-.05em{\sc i\kern-.025em b}\kern-.08em\TeX}

\begin{document}

%%% The following commands remove the headers in your paper. For final 
%%% papers, these will be inserted during the pagination process.

\pagestyle{fancy}
\fancyhead{}

% to print page numbers
\settopmatter{printfolios=true}

%%% The next command prints the information defined in the preamble.
\maketitle 
%%%%%%%%%%%%%%%%%%%%%%%%%%%%%%%%%%%%
%%%%% COMMENTED OUT         %%%%%
%%%%% TO AVOID PAGE NUMBER  %%%%%
%\thispagestyle{plain}
%\pagestyle{plain}

%%%%%%%%%%%%%%%%%%%%%%%%%%%%%%%%%%%%

%%%%%%%%%%%%%%%%%%%%%%%%%%%%%%%%%%%%%%%%%%%%%%%%%%%%%%%%%%%%%%%%%%%%%%%%

\section{Introduction}
Reinforcement Learning (RL) aims at the automatic acquisition of skills or some other 
form of intelligence, to behave appropriately and wisely in comparable situations and 
potentially on situations that are slightly different from the ones seen in training. 
When it comes to real world situations, there are two challenges: 
having a data-efficient learning method and being able to handle complex and unknown 
dynamical systems that can be difficult to model and are too far away from the systems
observed during the training phase. Because the dynamic nature of the 
environment may be challenging to learn, a first stream of RL methods has consisted 
in modeling the environment with a model. Hence it is called model-based RL. Model-based 
methods tend to excel in learning complex environments like financial markets. In 
mainstream agents literature, examples include, robotics applications, where it is 
highly desirable to learn using the lowest possible number of real-world trails 
\cite{Kaelbling_1996}. It is also used in finance where there are a lot of regime 
changes \cite{Freitas_2009,Niaki2013,Heaton_2017}. A first generation of model-based 
RL, relying on Gaussian processes and time-varying linear dynamical systems, provides 
excellent performance in low-data regimes \cite{deisenroth2011learning, deisenroth2011pilco, 
deisenroth2014gaussian,Levine_Koltun_2013,Kumar_2016}. A second generation, leveraging 
deep networks \cite{Gal_2016,Depeweg_2016,Nagabandi_2018}, has emerged and is based 
on the fact that neural networks offer high-capacity function approximators even 
in domains with high-dimensional observations \cite{Oh_2015,Ebert_2018,Kaiser_2019} 
while retaining some sample efficiency of a model-based approach. Recently, it has 
been proposed to adapt model-based RL via meta policy optimization to achieve asymptotic  
performance of model-free models \cite{Clavera_2018}. For a full survey of model-based 
RL model, please refer to \cite{moerland2020modelbased}. In finance, it is common 
to scale portfolio's allocations based on volatility and correlation as volatility 
is known to be a good proxy for the level of risk and correlation a standard measure 
of dependence. It is usually referred as volatility targeting. It enables the portfolio 
under consideration to achieve close to constant volatility through various market 
dynamics or regimes by simply sizing the portfolio's constituents according to volatility 
and correlation forecasts.

In contrast, the model-free approach aims to learn the optimal actions blindly without 
a representation of the environment dynamics. Works like \cite{Mnih_2015,Lillicrap_2016,Haarnoja_2018} 
have come with the promise that such models learn from raw inputs (and raw pixels) 
regardless of the game and provide some exciting capacities to handle new situations 
and environments, though at the cost of data efficiency as they require millions 
of training runs.

Hence, it is not surprising that the research community has focused on a new generation 
of models combining model-free and model-based RL approaches. A first idea has been to combine model-based and model-free updates for Trajectory-Centric 
RL. \cite{Chebotar_2017}. Another idea has been to use temporal difference models to have a model-free deep RL 
approach for model-based control \cite{Pong_2018}. \cite{van_Hasselt_2018} answers the question of  when to use parametric models in reinforcement learning. Likewise, \cite{Janner_2019} gives some hints when to trust model-based policy optimization versus model-free. 
\cite{Feinberg_2018} shows how to use model-based value estimation for efficient 
model-free RL.

All these studies, mostly applied to robotics and virtual environments, have not 
hitherto been widely used for financial time series. Our aim is to be able to distinguish 
various financial models that can be read or interpreted as model-based RL methods. 
These models aim at predicting volatility in financial markets in the context of 
portfolio allocation according to volatility target methods. These models are quite 
diverse and encompass statistical models based on historical data such as simple 
and naive moving average models, multivariate generalized auto-regressive conditional 
heteroskedasticity (GARCH) models, high-frequency based volatility models (HEAVY) 
\cite{Noureldin12multivariatehigh-frequency-based} and forward-looking models such 
as implied volatility or PCA decomposition of implied volatility indices. To be able 
to decide on an allocation between these various models, we rely on deep model-free 
RL. However, using just the last data points does not work in our cases as the various 
volatility models have very similar behaviors. Following \cite{Benhamou2020bridging} 
and \cite{Benhamou2021knowledge}, we also add contextual information like macro signals 
and risk appetite indices to include additional information in our DRL agent hereby 
allowing us to choose the pre-trained models that are best suited for a given environment.

\subsection{Related works}
The literature on portfolio allocation in finance using either supervised or reinforcement 
learning has been attracting more attention recently. Initially, \cite{Freitas_2009,Niaki2013,Heaton_2017} 
use deep networks to forecast next period prices and to use this prediction to infer 
portfolio allocations. The challenge of this approach is the weakness of predictions: 
financial markets are well known to be non-stationary and to present regime changes 
(see \cite{Salhi_2016,Dias_2015,benhamou2018trend,Zheng_2019}).

More recently, \cite{Jiang_2016,Zhengyao_2017,Liang_2018,Yu_2019,Wang_2019,Liu_2020,Ye_2020,Li_2019,Xiong_2018,Benhamou2020detecting,Benhamou2020time,Benhamou2021knowledge,Benhamou2020bridging} 
have started using deep reinforcement learning to do portfolio allocation. Transaction 
costs can be easily included in the rules. However, these studies rely on very distinct 
time series, which is a very different setup from our specific problem. They do not 
combine a model-based with a model-free approach. In addition, they do not investigate 
how to rank features, which is a great advantage of methods in ML like decision trees. 
Last but not least, they never test the statistical difference between the benchmark 
and the resulting model.

\subsection{Contribution}
Our contributions are precisely motivated by the shortcomings presented in the aforementioned 
remarks. They are four-fold:
\begin{itemize}
\item \textbf{The use of model-free RL to select various models that can be interpreted 
as model-based RL}. In a noisy and regime-changing environment like financial time 
series, the practitioners' approach is to use a model to represent the dynamics of 
financial markets. We use a model-free approach to learn from states to actions and 
hence distinguish between these initial models and choose which model-based RL to 
favor. In order to augment states, we use additional contextual information.

\item \textbf{The walk-forward procedure.} Because of the non stationary nature of 
time-dependent data, and especially financial data, it is crucial to test DRL model 
stability. We present a traditional methodology in finance but never used to our 
knowledge in DRL model evaluation, referred to as walk-forward analysis that iteratively 
trains and tests models on extending data sets. This can be seen as the analogy of 
cross-validation for time series. This allows us to validate that the selected hyper-parameters 
work well over time and that the resulting models are stable over time. 

\item \textbf{Features sensitivity procedure.} Inspired by the concept of feature 
importance in gradient boosting methods, we have created a feature importance of 
our deep RL model based on its sensitivity to features inputs. This allows us to 
rank each feature at each date to provide some explanations why our DRL agent chooses 
a particular action.

\item \textbf{A statistical approach to test model stability.} Most RL papers do 
not address the statistical difference between the obtained actions and predefined 
baselines or benchmarks. We introduce the concept of statistical difference as we 
want to validate that the resulting model is statistically different from the baseline 
results.
\end{itemize}

\section{Problem formulation}
Asset allocation is a major question for the asset management industry. It aims at 
finding the best investment strategy to balance risk versus reward by adjusting the 
percentage invested in each portfolio's asset according to risk tolerance, investment 
goals and horizons. 

Among these strategies, volatility targeting is very common. Volatility targeting 
forecasts the amount to invest in various assets based on their level of risk to 
target a constant and specific level of volatility over time. Volatility acts as 
a proxy for risk. Volatility targeting relies on the empirical evidence that a constant 
level of volatility delivers some added value in terms of higher returns and lower 
risk materialized by higher Sharpe ratios and lower drawdowns, compared to a buy 
and hold strategy \cite{Hocquard_2013,Perchet_2016,Dreyer_2017}. 
Indeed it can be shown that Sharpe ratio makes a lot of sense for manager to measure their performance. 
The distribution of Sharpe ratio can be computed explicitly \cite{benhamou2019connecting}. 
Sharpe ratio is not an accident and is a good indicator of manager performance \cite{benhamou2019testing}. 
It can also be related to other performance measures like Omega ratio \cite{benhamou2019omega} 
and other performance ratios \cite{benhamou2018incremental}. It also relies on 
the fact that past volatility largely predicts future near-term volatility, while 
past returns do not predict future returns. Hence, volatility is persistent, meaning 
that high and low volatility regimes tend to be followed by similar high and low 
volatility regimes. This evidence can be found not only in stocks, but also in bonds, 
commodities and currencies.  Hence, a common model-based RL approach for solving 
the asset allocation question is to model the dynamics of the future volatility. 

To articulate the problem, volatility is defined as the standard deviation of the 
returns of an asset. Predicting volatility can be done in multiple ways:
\begin{itemize}
\item Moving average: this model predicts volatility based on moving averages.
\item Level shift: this model is based on a two-step approach that allows the creation 
of abrupt jumps, another stylized fact of volatility. 
\item GARCH: a generalized auto-regressive conditional heteroske-dasticity model 
assumes that the return $r_t$ can be modeled by a time series $r_t = \mu + \epsilon_t$ 
where $\mu$ is the expected return 
and $\epsilon_t$ is a zero-mean white noise, and  $\epsilon_t = \sigma_t z_t$,
where $\sigma^2_t = \omega + \alpha  \epsilon^2_{t-1} + \beta \sigma^2_{t-1}$. The 
parameters $(\mu,\omega,\alpha,\beta)$ are estimated simultaneously by maximizing 
the log-likelihood.
\item GJR-GARCH: the Glosten-Jagannathan-Runkle GARCH (GJR-GARCH) model is a variation 
of the GARCH model (see \cite{Glosten_1993}) with the difference that $\sigma_t$, 
the  variance of the white noise  $\epsilon_t$,  is modelled as: 
$\sigma^2_t = \omega +(\alpha + \gamma_{t-1}) \epsilon^2_{t-1}+ \beta \sigma^2_{t-1}$ 
where $I_{t-1}=1$ if $r_{t-1}<\mu$ 
and 0 otherwise. The parameters $(\mu,\omega,\alpha,\gamma,\beta)$ 
are estimated simultaneously by maximizing the log-likelihood.
\item HEAVY: the HEAVY model utilizes high-frequency data for the objective of multi-step 
volatility forecasting
\cite{Noureldin12multivariatehigh-frequency-based}.
\item HAR: this model is an heterogeneous auto-regressive (HAR) model that aims at 
replicating how information actually flows in financial markets from long-term to 
short-term investors.
\item Adjusted TYVIX: this model uses the TYVIX index to forecast volatility in the 
bond future market,
\item Adjusted Principal Component: this model uses Principal Component Analysis 
to decompose a set of implied volatility indices into its main eigenvectors and renormalizes 
the resulting volatility proxy to match a realized volatilty metric.
\item RM2006: RM2006 uses a volatility forecast derived from an exponentially weighted 
moving average (EWMA) metric.
\end{itemize}

\begin{figure}[H]
    \centering
    \includegraphics[width= \linewidth, height=4cm]{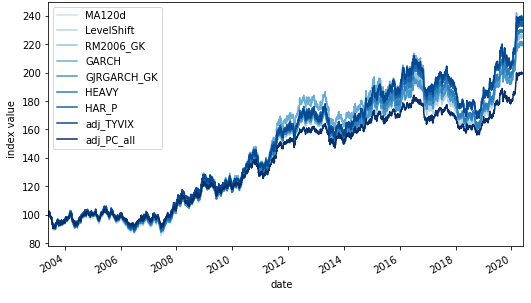}
    \caption{Volatility targeting model price evolution}
    \label{fig:model_evolution}
\end{figure}

\subsection{Mathematical formulation}
We have $n=9$ models. Each model predicts a volatility for the rolled U.S. 10-year 
note future contract that we shall call "bond future" in the remainder of this paper. 
The bond future's daily returns are denoted by $r^{bond}_t$ and typically range from -2 
to 2 percents with a daily average value of a few basis points and a daily standard 
deviation 10 to 50 times higher and ranging from 20 to 70 basis points. By these standards, 
the bond future's market is hard to predict and has a lot of noise making its forecast a
difficult exercise. Hence, using some volatility forecast to scale position makes a lot of sense.
These forecasts are then used to compute the allocation to the bond future's models. Mathematically, 
if the target volatility of the strategy is denoted by $\sigma_{target}$ and if the 
model $i$ predicts a bond future's volatility $\sigma^{i,pred}_{t-1}$, based on information 
up to $t-1$, the allocation in the bond future's model $i$  at time $t$ is given 
by the ratio between the target volatility and the predicted volatility: $k^i_{t-1} 
= \frac{\sigma_{target}}{\sigma^{i,pred}_{t-1}}$. \\

Hence, we can compute the daily amounts invested in each of the bond future volatility 
models and create a corresponding time series of returns $r^i_t = k^i_{t-1} \times 
r^{bond}_t$, consisting of investing in the bond future according to the allocation 
computed by the volatility targeting model $i$. This provides $n$ time series of 
compounded returns whose values are given by $P^i_t = \prod_{u=t_1...t}\left( 1+r^i_u 
\right)$. Our RL problem then boils down to selecting the optimal portfolio allocation 
(with respect to the cumulative reward) in each model-based RL strategies $a^i_t$ 
such that the portfolio weights sum up to one and are non-negative $\sum_{i=1..n} 
a^i_t=1$ and $a^i_t \geq 0$ for any $i=1..n$. These allocations are precisely the 
continuous actions of the DRL model. This is not an easy problem as the different 
volatility forecasts are quite similar. Hence, the $n$ time series of compounded 
returns look almost the same, making this RL problem non-trivial. Our aim is, in 
a sense, to distinguish between the indistinguishable strategies that are presented 
in figure \ref{fig:model_evolution}. More precisely, figure \ref{fig:model_evolution}
provides the evolution of the net value of an investment strategy that follows 
the different volatility targeting models.

\begin{figure}[H]
    \centering
    \includegraphics[width= \linewidth, height=4.5cm ]{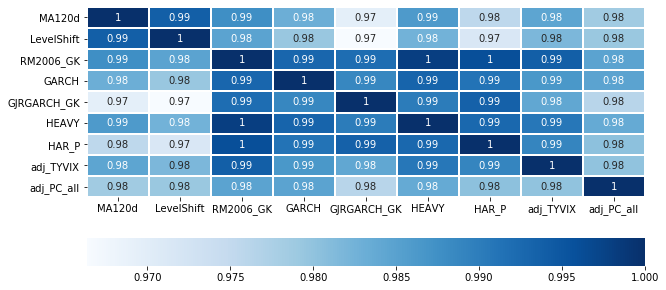}
    \caption{Correlation between the different volatility targeting models' returns}
    \label{fig:correlation}
\end{figure}

Compared to standard portfolio allocation problems, these strategies' returns are 
highly correlated and similar as presented by the correlation matrix \ref{fig:correlation}, 
with a lowest correlation of 97\%. The correlation is computed as the Pearson correlation over the full data set from 2004 to 2020.

Following \cite{SuttonBarto_2018}, we formulate this RL problem as a Markov Decision 
Process (MDP) problem. We define our MDP with a 6-tuple $\mathcal{M} = (T, \gamma, 
\mathcal{S}, \mathcal{A}, P, r)$
where  $T$ is the (possibly infinite) decision horizon, $\gamma \in \left]0, 1\right]$ the discount factor,  $\mathcal{S}$ the state space, $\mathcal{A}$ the action space, $p(s_{t+1}| s_{t}, a_{t})$  the transition probability from the state $s_{t}$ to $s_{t+1}$ given that the agent has chosen the action $a_t$, and $r(s_{t}, a_t)$ the reward for a state $s_{t}$ and an action $a_t$.

The agent's objective is to maximize its expected cumulative returns, given the start 
of the distribution. If we denote by $\pi$ the policy mapping specifying the action 
to choose in a particular state, $\pi : 
\mathcal{S} \to \mathcal{A}$, the agent wants to maximize the expected cumulative 
returns. This is written as: 
$J^{\pi} = \mathbb{E}_{s_t \sim P, a_t \sim \pi}\left[ \sum_{t=1}^{T} \gamma^{t-1} 
r(  s_{t}, a_t) \right]$. 

MDP assumes that we know all the states of the environment and have all the information 
to make the optimal decision in every state. 

From a practical standpoint, there are a few limitations to accommodate. First of 
all, the Markov property implies that knowing the current state is sufficient. Hence, 
we modify the RL setting by taking a pseudo state formed with a set of past observations 
$(o_{t-n}, o_{t-n-1}, \ldots, o_{t-1}, o_t)$. The trade-off is to take enough past 
observations to be close to a Markovian status without taking too many observations 
which would result in noisy states. 

In our settings, the actions are continuous and consist in finding at time $t$ the 
portfolio allocations $a^i_t$ in each volatility targeting model. We denote by $a_t=\left( 
a^1_t, ..., a^n_t\right)^T$ 
the portfolio weights vector. 

\begin{strip}
    \centering
    \includegraphics[width=\linewidth]{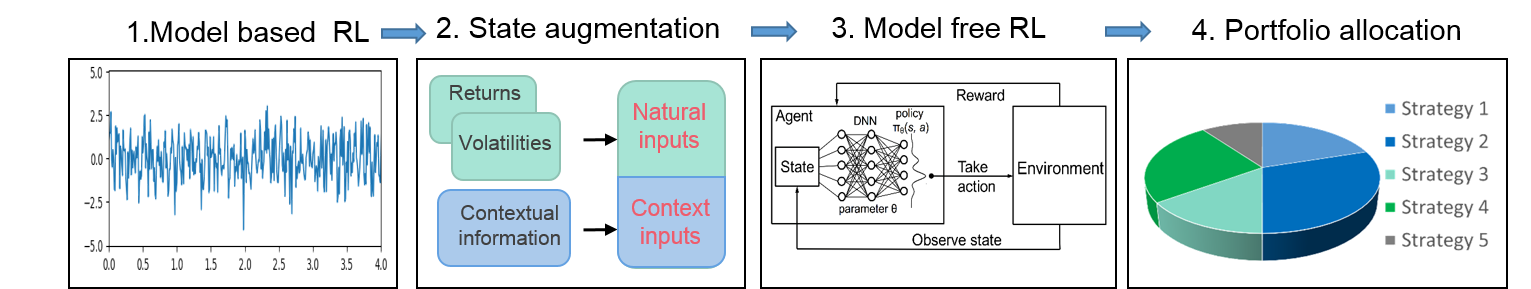}
    \captionsetup{type=figure}
    \captionof{figure}{Overall architecture}
    \label{fig:architecture}
\end{strip}

Likewise, we denote by $p_t=\left( p^1_t, ...,p^n_t\right)^T$ the closing price vector, 
and by 
$u_t = p_t  \oslash p_{t-1} = \left( p^1_t / p^1_{t-1}, ...,p^n_t / p^n_{t-1} \right)^T$ 
the price relative difference vector, where $\oslash$ denotes the element-wise division, 

and by $r_t=  \left( p^1_t / p^1_{t-1} - 1, ...,p^n_t / p^n_{t-1} -1 \right)^T$ the 
returns vector which is  also the percentage  change of each closing prices $p^1_t, 
...,p^n_t$.  Due to price change in the market, at the end of the same period, the 
weights evolve according to $w_{t-1} = (u_{t-1} \odot a_{t-1}) / (u_{t-1} . a_{t-1}) 
$ where $\odot$ is the element-wise multiplication, and $.$ the scalar product.

The goal of the agent at time $t$ is hence to reallocate the portfolio vector from 
$w^{t-1} $ to $a_t$ by buying and selling the relevant assets, taking into account 
the transaction costs that are given by
 $\alpha \lvert a_{t} - w_{t-1} \rvert_{1} $ 
 where $\alpha$ is the percentage cost for a transaction (which is quite low for 
future markets and given by 1 basis point) and $\lvert  .  | _{1} $ 
 is the $L_1$ norm operator. Hence at the end of time $t$, the agent receives a portfolio 
return given by $a_{t}. u_{t} -  \alpha \lvert   a_{t} - w_{t-1} \rvert_{1}$. 
 The cumulative reward corresponds to the sum of the logarithmic returns of the portfolio 
strategy given by 
 $\mathbb{E}\left[ \prod_{t=1}^T \log \left( a_{t}. u_{t} -  \alpha  \lvert a_{t} 
- w_{t-1} \rvert_{1} \right ) \right] $, 
 which is easier to process in a tensor flow graph as a log sum expression and is 
naturally given by $\mathbb{E}\left[ \log  \left(  \sum_{t=1}^T a_{t}. u_{t} -  \alpha 
|a_{t} - w_{t-1}|_{1} \right ) \right]$.

Actions are modeled by a multi-input, multi-layer convolution network whose details 
are given by Figure \ref{fig:network}. It has been shown that convolution networks are better for selecting features in portfolio allocation problem
\cite{Benhamou_DRPLECML}, \cite{Benhamou2020detecting} and \cite{Benhamou2020time}. The goal of the model-free RL method is to 
find the network parameters. This is done by an adversarial policy gradient method 
summarized by the algorithm \ref{alg1} using traditional Adam optimization so that 
we have the benefit of adaptive gradient descent with root mean square propagation 
\cite{kingma2014method} with a learning rate of 1\% and a number of iteration steps 
of 100,000 with an early stop criterion if the cumulative reward does not improve 
after 15 full episodes. Because each episode is run on the same financial data, we 
use on purpose a vanilla policy gradient algorithm to take advantage of the stability 
of the environment rather than to use more advanced DRL agents like TRPO, DDPG or 
TD3 that would add on top of our model free RL layer some extra complexity and noise.

 \begin{algorithm}[!htbp]
    \caption{Adversarial Policy Gradient}
    \label{alg1}
\begin{algorithmic}[1]
    \STATE Input: initial policy parameters $\theta$, empty replay buffer $\mathcal{D}$

\REPEAT
    \STATE Reset replay buffer
    \WHILE{not Terminal}
        \STATE Observe observation $o$ and select action $a = \pi_{\theta}(o)$ with 
probability $p$ and random action with probability $1-p$, 
        \STATE Execute $a$ in the environment    
        \STATE Observe next observation $o'$, reward $r$, and done signal $d$ to 
indicate whether $o'$ is terminal
        \STATE Apply noise to next observation $o'$
        \STATE Store $(o,a,o')$ in replay buffer $\mathcal{D}$
        \IF{Terminal}
            \FOR{however many updates in $\mathcal{D}$}
                \STATE Compute final reward $R$
            \ENDFOR
            \STATE Update network parameter with Adam gradient ascent
                $\vec\theta \longrightarrow \vec\theta + \lambda\nabla_{\vec\theta}J_{[0,t]}(\pi_{\vec 
\theta})$
        \ENDIF
    \ENDWHILE
\UNTIL{Convergence}
\end{algorithmic}
\end{algorithm}

\subsection{Benchmarks}
\subsubsection{Markowitz}
In order to benchmark our DRL approach, we need to compare to traditional financial 
methods. Markowitz allocation as presented in \cite{Markowitz_1952} is a widely-used 
benchmark in portfolio allocation as it is a straightforward and intuitive mix between 
performance and risk. In this approach, risk is represented by the variance of the 
portfolio. Hence, the Markowitz portfolio minimizes variance for a given expected 
return, which is solved by standard quadratic programming optimization. If we denote 
by $\mu = ( \mu_1, ..., \mu_n)^T$ the expected returns for our $n$ model strategies 
and by $\Sigma$ the covariance matrix of these strategies' returns, and by $r_{min}$ 
the targeted minimum return, the Markowitz optimization problem reads

\begin{eqnarray*}
&\text{Minimize} & w^T \Sigma w     \\
&\text{subject to}& \mu^T w \geq r_{min}  ,  \sum_{i=1 \ldots l} w_i = 1, w \geq 
0 
\end{eqnarray*}

\subsubsection{Average}
Another classical benchmark model for indistinguishable strategies, is the arithmetic 
average of all the volatility targeting models. This seemingly naive benchmark is 
indeed performing quite well as it mixes diversification 
and mean reversion effects.

\subsubsection{Follow the winner}
Another common strategy is to select the best performer of the past year, and use 
it the subsequent year. It replicates the standard investor's behavior 
that selects strategies that have performed well in the past. 

\subsection{Procedure and walk forward analysis}
The whole procedure is summarized by Figure \ref{fig:architecture}. We have $n$ models 
that represent the dynamics of the market volatility. We then add the volatility 
and the contextual information to the states, thereby yielding augmented states. 
The latter procedure is presented as the second step of the process. We then use 
a model-free RL approach to find the portfolio allocation among the various volatility 
targeting models, corresponding to steps 3 and 4. 
In order to test the robustness of our resulting DRL model, we introduce a new methodology 
called walk forward analysis.

\subsubsection{Walk forward analysis} In machine learning, the standard approach 
is to do $k$-fold cross-validation. This approach breaks the chronology of data and 
potentially uses past data in the test set. Rather, we can take sliding test set 
and take past data as training data. To ensure some stability, we favor to add incrementally 
new data in the training set, at each new step.

\vspace{0.1cm}
\begin{strip}
    \centering
    \includegraphics[width=\linewidth]{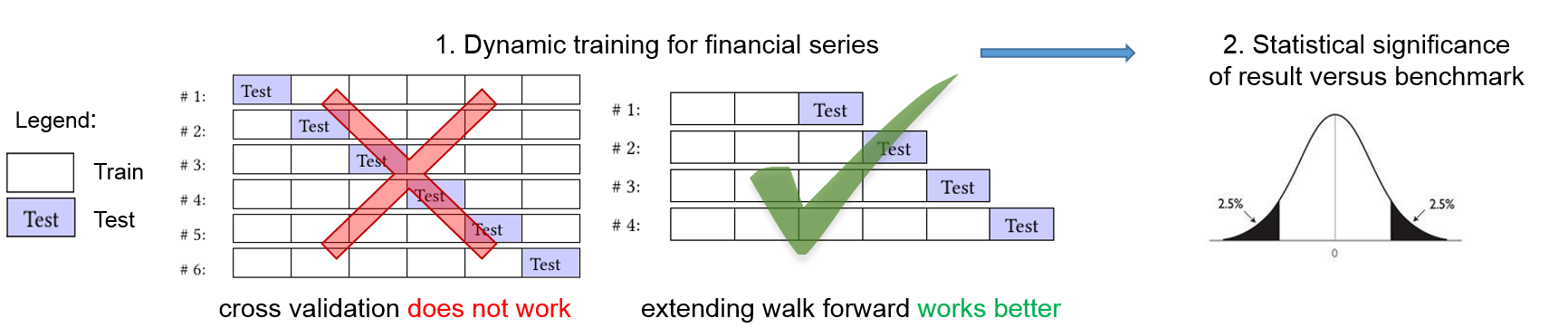}
    \captionsetup{type=figure}
    \captionof{figure}{Overall training process}
    \label{fig:process}
\end{strip}

 This method is sometimes referred to as "anchored walk forward" as we have anchored 
training data. Finally, as the test set is always after the training set, 
walk forward analysis gives less steps compared with cross-validation. In practice, 
and for our given data set, we train our models from 2000 to the end of 2013 (giving 
us at least 14 years of data) and use a repetitive test period of one year from 2014 
onward. Once a model has been selected, we also test its statistical significance, 
defined as the difference between the returns of two time series. We therefore do 
a T-test to validate how different these time series are. The whole process is summarized 
by Figure \ref{fig:process}. 
 
\subsubsection{Model architecture}

\begin{figure}[!htbp]
    \centering
    \includegraphics[width=\linewidth]{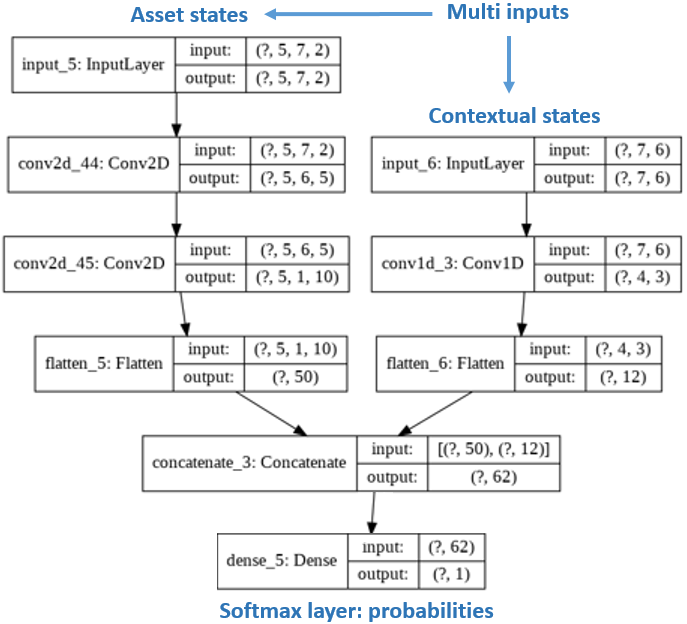}
    \caption{Multi-input DRL network}
    \label{fig:network}
\end{figure}

The states consist in two different types of data: the asset inputs and the contextual 
inputs.

Asset inputs are a truncated portion of the time series of financial returns 
of the volatility targeting models and of the volatility of these returns computed 
over a period of 20 observations. So if we denote by $r^i_t$ the returns of model 
$i$ at time $t$, and by $\sigma^i_{t}$ the standard deviation of returns over the 
last $d=20$ periods, asset inputs are given by a 3-D tensor denoted by $A_t = \left[ 
R_t, V_t\right]$, with 
\vspace{0.2cm}

\resizebox{0.9\linewidth} {!} {
$R_t =  \left( \!
\begin{array}{c}
r^1_{t-n} \,\,	... \,\, r^1_t \\  
... \,\,... \,\, ...\\ 
r^m_{t-n} \,\,.... \,\, r^m_t 
\end{array} \! \right)\! \  \ \\ \  
 \text{and  } V_t =  \left( \!
\begin{array}{c}
\sigma^1_{t-n} 	\,\,	... \,\, \sigma^1_t\\
... \,\,... \,\, ...\\
\sigma^m_{t-n} \,\,.... \,\, \sigma^m_t
\end{array} \! \right)$. }
\vspace{0.2cm}

This setting with two layers (past returns and past volatilities) is very different 
from the one presented in \citet{Jiang_2016,Zhengyao_2017,Liang_2018} that uses layers 
representing open, high, low and close prices, which are not necessarily available 
for volatility target models. Adding volatility is crucial to detect regime change 
and is surprisingly absent from these works.

Contextual inputs are a truncated portion of the time series of additional 
data that represent contextual information. Contextual information enables our DRL 
agent to learn the context, and are, in our problem, short-term and long-term risk 
appetite indices and short-term and long-term macro signals. Additionally, we include 
the maximum and minimum portfolio strategies return and the maximum portfolio strategies 
volatility. Similarly to asset inputs, standard deviations is useful to detect regime 
changes. Contextual observations are stored in a 2D matrix denoted by $C_t$ with 
stacked past $p$ individual contextual observations. The contextual state reads

$
C_t =  \left( \!\!
\begin{array}{c  }
c^1_{t-n} 	\,\,	... \,\, c^1_t\\
... \,\,... \,\, ...\\
c^p_{t-n} 	\,\,.... \,\, c^p_t
\end{array} \!\!\! \right)$. 

The output of the network is a softmax layer that provides the various allocations. 
As the dimensions of the assets and the contextual inputs are different, the network 
is a multi-input network with various convolutional layers and a final softmax dense 
layer as represented in Figure \ref{fig:network}. 

\subsubsection{Features sensitivity analysis}
One of the challenges of neural networks relies in the difficulty to provide explainability 
about their behaviors. Inspired by computer vision, we present a methodology here 
that enables us to relate features to action. This concept is based on features sensitivity 
analysis. Simply speaking, our neural network is a multi-variate function. Its inputs 
include all our features, strategies, historical performances, standard deviations, 
contextual information, short-term and long-term macro signals and risk appetite 
indices. We denote these inputs by $X$, which lives in
$\mathbb{R}^k$ where $k$ is the number of features.
Its outputs are the action vector $Y$, which is an $n$-d array with elements between 
0 and 1. This action vector lives in an image set denoted by $\mathcal{Y}$, which 
is a subset of $\mathbb{R}^n$ . Hence, the neural network is a function $\Phi: \mathbb{R}^k 
\to \mathcal{Y}$ with $\Phi(X)=Y$. In order to project the various partial derivatives, 
we take the L1 norm (denoted by $|.|_{1}$) of the different partial derivatives as 
follows:
$ |\frac{ \partial  \Phi(X) } {\partial X}|_{1}$. The choice of the L1 norm is arbitrary 
but is intuitively motivated by the fact that we want to scale the distance of the 
gradient linearly.

%\begin{figure}[H]
%    \centering
%    \includegraphics[width=\linewidth]{images/excited_neurons.png}
%    \caption{Features sensitivity summary}
%    \label{fig:excited_neurons}
%\end{figure}

In order to measure the sensitivity of the outputs, simply speaking, we change the 
initial feature by its mean value over the last $d$ periods. This is inspired by 
a "what if" analysis where we would like to measure the impact of changing the feature 
from its mean value to its current value. In computer vision, the practice is not 
to use the mean value but rather to switch off the pixel and set it to the black 
pixel. In our case, using a zero value would not be relevant as this would favor 
large features. We are really interested here in measuring the sensitivity of our 
actions when a feature deviates from its mean value.

\begin{figure}[H]
    \centering
    \includegraphics[width=\linewidth]{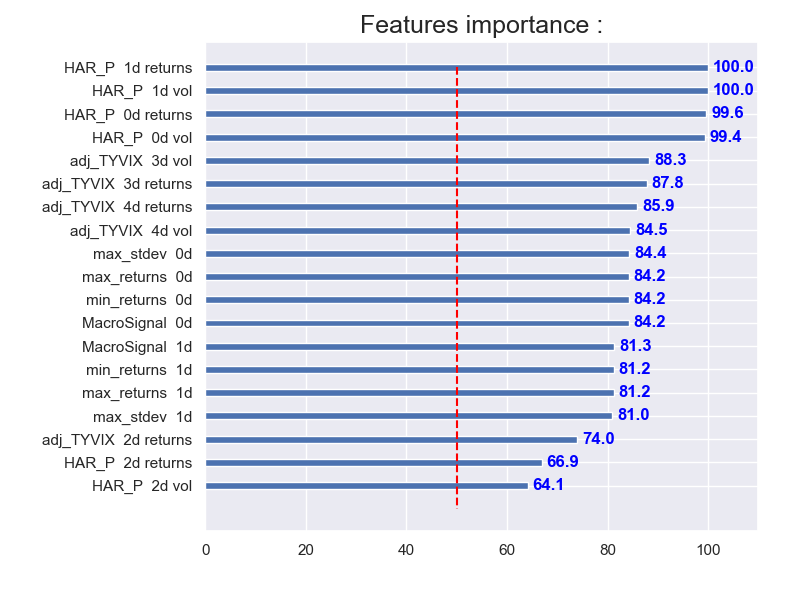}
    \caption{Model explainability}
    \label{fig:explainability}
\end{figure}

The resulting value is computed numerically and provides us for each feature a feature 
importance. We rank these features importance and assign arbitrarily the value $100$ 
to the largest and $0$ to the lowest. 
This provides us with the following features importance plot given below \ref{fig:explainability}. 
We can notice that the HAR returns and volatility are the most important features, 
followed by various returns and volatility for the TYVIX model. Although returns 
and volatility are dominating among the most important features, macro signals 0d 
observations comes as the 12th most important feature over 70 features with a very 
high score of 84.2. The features sensitivity analysis confirms two things: i) it 
is useful to include volatility features as they are good predictors of regime changes, 
ii) contextual information plays a role as illustrated by the macro signal.

\section{Out of sample results}
\begin{figure}[htbp]
    \centering
    \includegraphics[width=\linewidth]{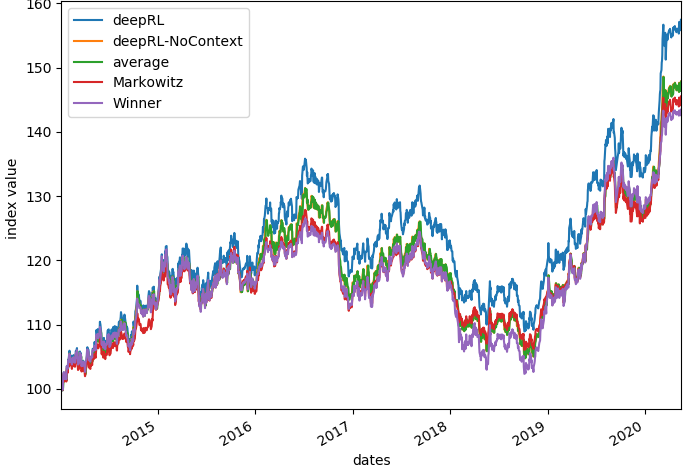}
    \caption{Model comparison}
    \label{fig:modelComparison}
\end{figure}

In this section, we compare the various models: the deep RL model (DRL1) using states 
with contextual inputs and standard deviation, the deep RL model without contextual 
inputs and standard deviation (DRL2), the average strategy, the Markowitz portfolio 
and the "the winner" strategy. The results are the combination of the 7 distinct 
test periods: each year from 2014 to 2020. The resulting performance is plotted in 
Figure \ref{fig:modelComparison}. We notice that the deepRL model with contextual 
information and standard deviation is substantially higher than the other models 
in terms of performance as it ends at 157, whereas other models (the deepRL with 
no context, the average, the Markowitz and "the winner" model) end at 147.6, 147.8, 
145.5, 143.4  respectively.

\begin{figure}[htbp]
    \centering
    \includegraphics[width=\linewidth, height=4.5cm]{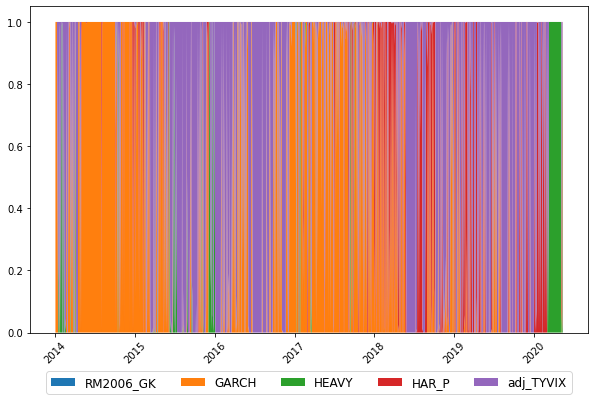}
    \caption{DRL portfolio allocation}
    \label{fig:weights}
\end{figure}

\begin{figure}[htbp]
    \centering
    \includegraphics[width=\linewidth]{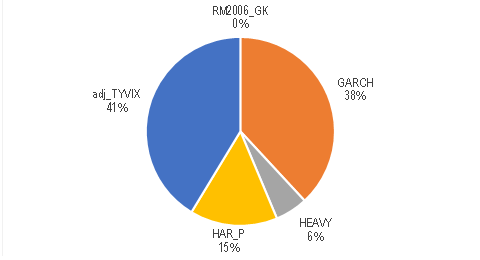}
    \caption{Average model allocation}
    \label{fig:average_model_allocation}
\end{figure}

To make such a performance, the DRL model needs to frequently rebalance between the 
various models (Figure \ref{fig:weights}) with  dominant allocations in GARCH and 
TYVIX models (Figure \ref{fig:average_model_allocation}).

\subsection{Results description}
\subsubsection{Risk metrics}
We provide various statistics in Table \ref{tab:Model comparison} for different time 
horizons: 1, 3 and 5 years. For each horizon, we put the best model, according to 
the column's criterion, in bold. The Sharpe and Sortino ratios are computed on daily 
returns. Maximum drawdown (written as mdd in the table), which is the maximum observed 
loss from a peak to a trough for a portfolio, is also computed on daily returns. 
DRL1 is the DRL model with standard deviations and contextual information, while 
DRL2 is a model with no contextual information and no standard deviation. Overall, 
DLR1, the DRL model with contextual information and standard deviation, performs 
better for 1, 3 and 5 years except for three-year maximum drawdown. Globally, it 
provides a 1\% increase in annual net return for a 5-year horizon. It also increases 
the Sharpe ratio by 0.1 and is able to reduce most of the maximum drawdowns except 
for the 3-year period. Markowitz portfolio selection and "the winner" strategy, which 
are both traditional financial methods heavily used by practitioners, do not work 
that well compared with a naive arithmetic average and furthermore when compared 
to the DRL model with context and standard deviation inputs. A potential explanation 
may come from the fact that these volatility targeting strategies are very similar 
making the diversification effects non effective.

\begin{table}[htbp]
  \centering
  \caption{Models comparison over 1, 3, 5 years}\label{tab:Model comparison}
    \resizebox{\linewidth} {!} {
    \begin{tabular}{|l|rrrrr|}
    \toprule
          & \multicolumn{1}{l}{ return } & \multicolumn{1}{l}{ sharpe } & \multicolumn{1}{l}{ 
sortino } & \multicolumn{1}{l}{ mdd } & \multicolumn{1}{l|}{ mdd/vol } \\
    \midrule
          &       & & 1 Year        &       &  \\
    \midrule
 DRL1 & \textbf{ 22.659} & \textbf{ 2.169} & \textbf{ 2.419} & \textbf{- 6.416} & 
\textbf{ - 0.614} \\
 DRL2 & 20.712 & 2.014 & 2.167 & - 6.584 & - 0.640 \\
 Average & 20.639 & 2.012 & 2.166 & - 6.560 & - 0.639 \\
 Markowitz & 19.370 & 1.941 & 2.077 & - 6.819 & - 0.683 \\
 Winner & 17.838 & 1.910 & 2.062 & - 6.334 & - 0.678 \\
 \midrule
  & &  &3 Years  & &  \\ 
 \midrule
 DRL1 & \textbf{ 8.056} & \textbf{ 0.835} & \textbf{ 0.899} & - 17.247 & \textbf{ 
- 1.787} \\
 DRL2 & 7.308 & 0.783 & 0.834 & \textbf{- 16.912} & - 1.812 \\
 Average & 7.667 & 0.822 & 0.876 & - 16.882 & - 1.810 \\
 Markowitz & 7.228 & 0.828 & 0.891 & - 16.961 & - 1.869 \\
 Winner & 6.776 & 0.712 & 0.754 & - 17.770 & - 1.867 \\
 \midrule
 & & & 5 Years  & & \\
 \midrule
 DRL1 & \textbf{ 6.302} & \textbf{ 0.651} & \textbf{ 0.684} & \textbf{- 19.794} & 
\textbf{ - 2.044} \\
 DRL2 & 5.220 & 0.565 & 0.584 & - 20.211 & - 2.187 \\
 Average & 5.339 & 0.579 & 0.599 & - 20.168 & - 2.187 \\
 Markowitz & 4.947 & 0.569 & 0.587 & - 19.837 & - 2.074 \\
 Winner & 4.633 & 0.508 & 0.526 & - 19.818 & - 2.095 \\
    \bottomrule
    \end{tabular}%
    }
  \label{tab:model_comparison}%
\end{table}%

\subsubsection{Statistical significance}
Following our methodology described in \ref{fig:process}, once we have computed the 
results for the various walk-forward test periods, we do a T-statistic test to validate 
the significance of the result. Given two models, we test the null hypothesis that 
the difference of the returns running average (computed as $(\sum_{u=0}^t r_u / t)$ 
for various times $t$) between the two models is equal to 0. We provide the T-statistic 
and, in parenthesis, the p-value. We take a p-value threshold of 5\%, and put the 
cases where we can reject the null hypothesis in bold in table \ref{tab:TStatRunningAvgReturns}. 
Hence, we conclude that the DRL model with context (DRL1) model is statistically 
different from all other models. These results on the running average are quite intuitive 
as we are able to distinguish the DRL1 model curve from all other curves in Figure 
\ref{fig:modelComparison}. Interestingly, we can see that the DRL model without context 
(DRL2) is not statically different from a pure averaging of the average model that 
consists in averaging allocation computed by model-based RL approaches.

\begin{table}[htbp]
  \centering
  \caption{T-statistics and P-values (in parenthesis) for running average returns 
difference}
      \begin{tabular}{|l|r|r|r|r|}
    \toprule
    Avg Return & \multicolumn{1}{l|}{DRL2} & \multicolumn{1}{l|}{Average} & \multicolumn{1}{l|}{Markowitz} 
& \multicolumn{1}{l|}{Winner} \\
    \midrule
    DRL1  & \textbf{72.1 (0\%)} & \textbf{14 (0\%)} & \textbf{44.1 (0\%)} & \textbf{79.8 
(0\%)} \\
    \midrule
    DRL2  &       & 1.2 (22.3\%) & \textbf{24.6 (0\%)} & \textbf{10 (0\%)} \\
    \midrule
    Average &       &       & \textbf{7.6(0\%)} & 0.9 (38.7\%) \\
    \midrule
    Markowitz &       &       &       & \textbf{-13.1 (0\%)} \\
    \bottomrule
    \end{tabular}%
  \label{tab:TStatRunningAvgReturns}%
\end{table}%

\subsubsection{Results discussion}
\begin{figure}[H]
    \centering
    \includegraphics[width=\linewidth, height=4cm]{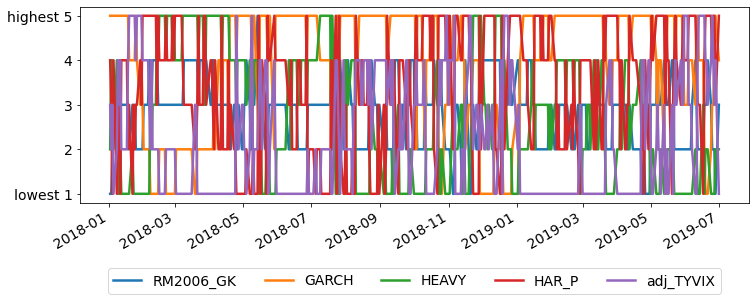}

    \caption{Volatility estimates rank}
    
    \label{fig:volatility_estimates_rank}
\end{figure}

It is interesting to understand how the DRL model achieves such a performance as 
it provides an amazing additional 1\% annual return over 5 years, and an increase 
in Sharpe ratio of 0.10. This is done simply by selecting the right strategies at 
the right time. This helps us to confirm that the adaptive learning thanks to the model free 
RL is somehow able to pick up regime changes. We notice that the DRL model selects 
the GARCH model quite often and, more recently, the HAR and HEAVY model (Figure \ref{fig:weights}). 
When targeting a given volatility level, capital weights are inversely proportional 
to the volatility estimates. Hence, lower volatility estimates give higher weights 
and in a bullish market give higher returns. Conversely, higher volatility estimates 
drive capital weights lower and have better performance in a bearish market. The 
allocation of these models evolve quite a bit as shown by Figure \ref{fig:volatility_estimates_rank}, 
which plots the rank of the first 5 models.

\begin{figure}[H]
    \centering
    \includegraphics[width=\linewidth, height=3.75cm]{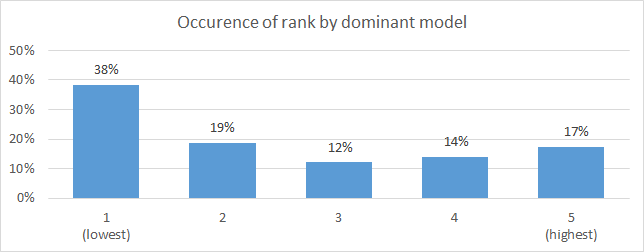}

    \caption{Occurrence of rank for the DRL model}
    \label{fig:occurence_rank}
\end{figure}

We can therefore test if the DRL model has a tendency to select volatility targeting 
models that favor lower volatility estimates. If we plot the occurrence of rank by 
dominant model for the DRL model, we observe that the DRL model selects the lowest 
volatility estimate model quite often (38.2\% of the time) but also tends to select 
the highest volatility models giving a U shape to the occurrence of rank as shown in figure \ref{fig:occurence_rank}. This U shape confirms two things: i) the model 
has a tendency to select either the lowest or highest volatility estimates models, 
which are known to perform best in bullish markets or bearish markets (however, it 
does not select these models blindly as it is able to time when to select the lowest 
or highest volatility estimates); ii) the DRL model is able to reduce maximum drawdowns 
while increasing net annual returns as seen in Table \ref{tab:Model comparison}. 
This capacity to simultaneously increase net annual returns and decrease maximum 
drawdowns indicates a capacity to detect regime changes. Indeed, a random guess would 
only increase the leverage when selecting lowest volatility estimates, thus resulting 
in higher maximum drawdowns.

\subsection{Benefits of DRL}
The advantages of context based DRL are numerous: (i) by design, DRL directly maps 
market conditions to actions and can thus adapt to regime changes, (ii) DRL can incorporate 
additional data and be a multi-input method, as opposed to more traditional optimization 
methods.

\subsection{Future work}
As nice as this may look, there is room for improvement as more contextual data and 
architectural networks choices could be tested as well as other DRL agents like DDPG, 
TRPO or TD3. It is also worth mentioning that the analysis has been conducted on a 
single financial instrument and a relatively short out-of-sample period. 
Expanding this analysis further in the past would cover more various regimes 
(recessions, inflationary, growth, etc.) and 
potentially improve the statistical relevance of this study at the cost of losing 
relevance for more recent data. Another lead consists of applying the same methodology 
to a much wider ensemble of securities and identify specific statistical features 
based on distinct geographic and asset sectors. 

\section{Conclusion}
In this work, we propose to create an adaptive learning method that combines model-based 
and model-free RL approaches to address volatility regime changes in financial markets. 
The model-based approach enables to capture efficiently the volatility dynamics while 
the model-free RL approach to time when to switch from one to another model. This 
combination enables us to have an adaptive agent that switches between different 
dynamics. We strengthen the model-free RL step with additional inputs like volatility 
and macro and risk appetite signals that act as contextual information. The ability 
of this method to reduce risk and profitability are verified when compared to the 
various financial benchmarks. The use of successive training and testing sets enables 
us to stress test the robustness of the resulting agent. Features sensitivity analysis 
confirms the importance of volatility and contextual variables and explains in part 
the DRL agent's better performance. Last but not least, statistical tests validate 
that results are statistically significant from a pure averaging method of all model-based 
RL allocations.

\bibliographystyle{ACM-Reference-Format} 
\balance
\bibliography{biblio}


\begin{thebibliography}{55}

%%% the Web address.

\ifx \showCODEN    \undefined \def \showCODEN     #1{\unskip}     \fi
\ifx \showDOI      \undefined \def \showDOI       #1{#1}\fi
\ifx \showISBNx    \undefined \def \showISBNx     #1{\unskip}     \fi
\ifx \showISBNxiii \undefined \def \showISBNxiii  #1{\unskip}     \fi
\ifx \showISSN     \undefined \def \showISSN      #1{\unskip}     \fi
\ifx \showLCCN     \undefined \def \showLCCN      #1{\unskip}     \fi
\ifx \shownote     \undefined \def \shownote      #1{#1}          \fi
\ifx \showarticletitle \undefined \def \showarticletitle #1{#1}   \fi
\ifx \showURL      \undefined \def \showURL       {\relax}        \fi
% invisible to TeX
\providecommand\bibfield[2]{#2}
\providecommand\bibinfo[2]{#2}
\providecommand\natexlab[1]{#1}
\providecommand\showeprint[2][]{arXiv:#2}

\bibitem[\protect\citeauthoryear{Benhamou}{Benhamou}{2018}]%
        {benhamou2018trend}
\bibfield{author}{\bibinfo{person}{Eric Benhamou}.}
  \bibinfo{year}{2018}\natexlab{}.
\newblock \showarticletitle{Trend without hiccups: a Kalman filter approach}.
\newblock \bibinfo{journal}{\emph{ssrn.1808.03297}} (\bibinfo{year}{2018}).
\newblock


\bibitem[\protect\citeauthoryear{Benhamou}{Benhamou}{2019}]%
        {benhamou2019connecting}
\bibfield{author}{\bibinfo{person}{Eric Benhamou}.}
  \bibinfo{year}{2019}\natexlab{}.
\newblock \showarticletitle{Connecting Sharpe ratio and Student t-statistic,
  and beyond}.
\newblock \bibinfo{journal}{\emph{ArXiv}} (\bibinfo{year}{2019}).
\newblock
\showeprint[arxiv]{1808.04233}~[q-fin.ST]


\bibitem[\protect\citeauthoryear{Benhamou and Guez}{Benhamou and Guez}{2018}]%
        {benhamou2018incremental}
\bibfield{author}{\bibinfo{person}{Eric Benhamou} {and}
  \bibinfo{person}{Beatrice Guez}.} \bibinfo{year}{2018}\natexlab{}.
\newblock \showarticletitle{Incremental Sharpe and other performance ratios}.
\newblock \bibinfo{journal}{\emph{Journal of Statistical and Econometric
  Methods}}  \bibinfo{volume}{2018} (\bibinfo{year}{2018}).
\newblock


\bibitem[\protect\citeauthoryear{Benhamou, Guez, and Paris1}{Benhamou
  et~al\mbox{.}}{2019a}]%
        {benhamou2019omega}
\bibfield{author}{\bibinfo{person}{Eric Benhamou}, \bibinfo{person}{Beatrice
  Guez}, {and} \bibinfo{person}{Nicolas Paris1}.}
  \bibinfo{year}{2019}\natexlab{a}.
\newblock \showarticletitle{Omega and Sharpe ratio}.
\newblock \bibinfo{journal}{\emph{ArXiv}} (\bibinfo{year}{2019}).
\newblock
\showeprint[arxiv]{1911.10254}~[q-fin.RM]


\bibitem[\protect\citeauthoryear{Benhamou, Saltiel, Guez, and Paris}{Benhamou
  et~al\mbox{.}}{2019b}]%
        {benhamou2019testing}
\bibfield{author}{\bibinfo{person}{Eric Benhamou}, \bibinfo{person}{David
  Saltiel}, \bibinfo{person}{Beatrice Guez}, {and} \bibinfo{person}{Nicolas
  Paris}.} \bibinfo{year}{2019}\natexlab{b}.
\newblock \showarticletitle{Testing Sharpe ratio: luck or skill?}
\newblock \bibinfo{journal}{\emph{ArXiv}} (\bibinfo{year}{2019}).
\newblock
\showeprint[arxiv]{1905.08042}~[q-fin.RM]


\bibitem[\protect\citeauthoryear{Benhamou, Saltiel, Ohana, and Atif}{Benhamou
  et~al\mbox{.}}{2021a}]%
        {Benhamou2020detecting}
\bibfield{author}{\bibinfo{person}{Eric Benhamou}, \bibinfo{person}{David
  Saltiel}, \bibinfo{person}{Jean-Jacques Ohana}, {and} \bibinfo{person}{Jamal
  Atif}.} \bibinfo{year}{2021}\natexlab{a}.
\newblock \showarticletitle{Detecting and adapting to crisis pattern with
  context based Deep Reinforcement Learning}. In
  \bibinfo{booktitle}{\emph{International Conference on Pattern Recognition
  (ICPR)}}. \bibinfo{publisher}{{IEEE} Computer Society}.
\newblock


\bibitem[\protect\citeauthoryear{Benhamou, Saltiel, Ohana, Atif, and
  Laraki}{Benhamou et~al\mbox{.}}{2021b}]%
        {Benhamou_DRPLECML}
\bibfield{author}{\bibinfo{person}{Eric Benhamou}, \bibinfo{person}{David
  Saltiel}, \bibinfo{person}{Jean~Jacques Ohana}, \bibinfo{person}{Jamal Atif},
  {and} \bibinfo{person}{Rida Laraki}.} \bibinfo{year}{2021}\natexlab{b}.
\newblock \showarticletitle{Deep Reinforcement Learning (DRL) for Portfolio
  Allocation}. In \bibinfo{booktitle}{\emph{Machine Learning and Knowledge
  Discovery in Databases. Applied Data Science and Demo Track}},
  \bibfield{editor}{\bibinfo{person}{Yuxiao Dong}, \bibinfo{person}{Georgiana
  Ifrim}, \bibinfo{person}{Dunja Mladeni{\'{c}}}, \bibinfo{person}{Craig
  Saunders}, {and} \bibinfo{person}{Sofie Van~Hoecke}} (Eds.).
  \bibinfo{publisher}{Springer International Publishing},
  \bibinfo{address}{Cham}, \bibinfo{pages}{527--531}.
\newblock


\bibitem[\protect\citeauthoryear{Benhamou, Saltiel, Ungari, and
  Abhishek~Mukhopadhyay}{Benhamou et~al\mbox{.}}{2021c}]%
        {Benhamou2021knowledge}
\bibfield{author}{\bibinfo{person}{Eric Benhamou}, \bibinfo{person}{David
  Saltiel}, \bibinfo{person}{Sandrine Ungari}, {and}
  \bibinfo{person}{Rida~Laraki Abhishek~Mukhopadhyay, Jamal~Atif}.}
  \bibinfo{year}{2021}\natexlab{c}.
\newblock \showarticletitle{Knowledge discovery with Deep RL for selecting
  financial hedges}. In \bibinfo{booktitle}{\emph{AAAI: KDF}}.
  \bibinfo{publisher}{{AAAI} Press}.
\newblock


\bibitem[\protect\citeauthoryear{Benhamou, Saltiel, Ungari, and
  Mukhopadhyay}{Benhamou et~al\mbox{.}}{2020a}]%
        {Benhamou2020bridging}
\bibfield{author}{\bibinfo{person}{Eric Benhamou}, \bibinfo{person}{David
  Saltiel}, \bibinfo{person}{Sandrine Ungari}, {and} \bibinfo{person}{Abhishek
  Mukhopadhyay}.} \bibinfo{year}{2020}\natexlab{a}.
\newblock \showarticletitle{Bridging the gap between Markowitz planning and
  deep reinforcement learning}. In \bibinfo{booktitle}{\emph{Proceedings of the
  30th International Conference on Automated Planning and Scheduling ({ICAPS}):
  PRL}}. \bibinfo{publisher}{{AAAI} Press}.
\newblock


\bibitem[\protect\citeauthoryear{Benhamou, Saltiel, Ungari, and
  Mukhopadhyay}{Benhamou et~al\mbox{.}}{2020b}]%
        {Benhamou2020time}
\bibfield{author}{\bibinfo{person}{Eric Benhamou}, \bibinfo{person}{David
  Saltiel}, \bibinfo{person}{Sandrine Ungari}, {and} \bibinfo{person}{Abhishek
  Mukhopadhyay}.} \bibinfo{year}{2020}\natexlab{b}.
\newblock \showarticletitle{Time your hedge with Deep Reinforcement Learning}.
  In \bibinfo{booktitle}{\emph{Proceedings of the 30th International Conference
  on Automated Planning and Scheduling ({ICAPS}): FinPlan}}.
  \bibinfo{publisher}{{AAAI} Press}.
\newblock


\bibitem[\protect\citeauthoryear{Chebotar, Hausman, Zhang, Sukhatme, Schaal,
  and Levine}{Chebotar et~al\mbox{.}}{2017}]%
        {Chebotar_2017}
\bibfield{author}{\bibinfo{person}{Yevgen Chebotar}, \bibinfo{person}{Karol
  Hausman}, \bibinfo{person}{Marvin Zhang}, \bibinfo{person}{Gaurav Sukhatme},
  \bibinfo{person}{Stefan Schaal}, {and} \bibinfo{person}{Sergey Levine}.}
  \bibinfo{year}{2017}\natexlab{}.
\newblock \showarticletitle{Combining Model-Based and Model-Free Updates for
  Trajectory-Centric Reinforcement Learning}. In
  \bibinfo{booktitle}{\emph{PMLR}} \emph{(\bibinfo{series}{Proceedings of
  Machine Learning Research})}, \bibfield{editor}{\bibinfo{person}{Doina
  Precup} {and} \bibinfo{person}{Yee~Whye Teh}} (Eds.),
  Vol.~\bibinfo{volume}{70}. \bibinfo{publisher}{PMLR},
  \bibinfo{address}{International Convention Centre, Sydney, Australia},
  \bibinfo{pages}{703--711}.
\newblock


\bibitem[\protect\citeauthoryear{Clavera, Rothfuss, Schulman, Fujita, Asfour,
  and Abbeel}{Clavera et~al\mbox{.}}{2018}]%
        {Clavera_2018}
\bibfield{author}{\bibinfo{person}{Ignasi Clavera}, \bibinfo{person}{Jonas
  Rothfuss}, \bibinfo{person}{John Schulman}, \bibinfo{person}{Yasuhiro
  Fujita}, \bibinfo{person}{Tamim Asfour}, {and} \bibinfo{person}{Pieter
  Abbeel}.} \bibinfo{year}{2018}\natexlab{}.
\newblock \bibinfo{title}{Model-Based Reinforcement Learning via Meta-Policy
  Optimization}.
\newblock
\newblock
\showeprint[arxiv]{1809.05214}
\urldef\tempurl%
\url{http://arxiv.org/abs/1809.05214}
\showURL{%
\tempurl}


\bibitem[\protect\citeauthoryear{Deisenroth, Fox, and Rasmussen}{Deisenroth
  et~al\mbox{.}}{2014}]%
        {deisenroth2014gaussian}
\bibfield{author}{\bibinfo{person}{M. Deisenroth}, \bibinfo{person}{D. Fox},
  {and} \bibinfo{person}{C. Rasmussen}.} \bibinfo{year}{2014}\natexlab{}.
\newblock \bibinfo{title}{{G}aussian processes for data-efficient learning in
  robotics and control}.
\newblock
\newblock


\bibitem[\protect\citeauthoryear{Deisenroth and Rasmussen}{Deisenroth and
  Rasmussen}{2011}]%
        {deisenroth2011pilco}
\bibfield{author}{\bibinfo{person}{Marc Deisenroth} {and}
  \bibinfo{person}{Carl~E Rasmussen}.} \bibinfo{year}{2011}\natexlab{}.
\newblock \bibinfo{title}{{PILCO}: A model-based and data-efficient approach to
  policy search}.
\newblock
\newblock


\bibitem[\protect\citeauthoryear{Deisenroth, Rasmussen, and Fox}{Deisenroth
  et~al\mbox{.}}{2011}]%
        {deisenroth2011learning}
\bibfield{author}{\bibinfo{person}{Marc~Peter Deisenroth},
  \bibinfo{person}{Carl~Edward Rasmussen}, {and} \bibinfo{person}{Dieter Fox}.}
  \bibinfo{year}{2011}\natexlab{}.
\newblock \bibinfo{title}{Learning to Control a Low-Cost Manipulator using
  Data-Efficient Reinforcement Learning.}
\newblock
\newblock
\showISBNx{978-0-262-51779-9}


\bibitem[\protect\citeauthoryear{Depeweg, Hern{\'a}ndez-Lobato, Doshi-Velez,
  and Udluft}{Depeweg et~al\mbox{.}}{2016}]%
        {Depeweg_2016}
\bibfield{author}{\bibinfo{person}{Stefan Depeweg},
  \bibinfo{person}{Jos{\'e}~Miguel Hern{\'a}ndez-Lobato},
  \bibinfo{person}{Finale Doshi-Velez}, {and} \bibinfo{person}{Steffen
  Udluft}.} \bibinfo{year}{2016}\natexlab{}.
\newblock \bibinfo{title}{Learning and policy search in stochastic dynamical
  systems with bayesian neural networks}.
\newblock
\newblock


\bibitem[\protect\citeauthoryear{Dias, Vermunt, and Ramos}{Dias
  et~al\mbox{.}}{2015}]%
        {Dias_2015}
\bibfield{author}{\bibinfo{person}{José Dias}, \bibinfo{person}{Jeroen
  Vermunt}, {and} \bibinfo{person}{Sofia Ramos}.}
  \bibinfo{year}{2015}\natexlab{}.
\newblock \showarticletitle{Clustering financial time series: New insights from
  an extended hidden Markov model}.
\newblock \bibinfo{journal}{\emph{European Journal of Operational Research}}
  \bibinfo{volume}{243} (\bibinfo{date}{06} \bibinfo{year}{2015}),
  \bibinfo{pages}{852--864}.
\newblock


\bibitem[\protect\citeauthoryear{Dreyer and Hubrich}{Dreyer and
  Hubrich}{2017}]%
        {Dreyer_2017}
\bibfield{author}{\bibinfo{person}{Anna Dreyer} {and} \bibinfo{person}{Stefan
  Hubrich}.} \bibinfo{year}{2017}\natexlab{}.
\newblock \bibinfo{title}{Tail Risk Mitigation with Managed Volatility
  Strategies}.
\newblock
\newblock


\bibitem[\protect\citeauthoryear{Ebert, Finn, Dasari, Xie, Lee, and
  Levine}{Ebert et~al\mbox{.}}{2018}]%
        {Ebert_2018}
\bibfield{author}{\bibinfo{person}{Frederik Ebert}, \bibinfo{person}{Chelsea
  Finn}, \bibinfo{person}{Sudeep Dasari}, \bibinfo{person}{Annie Xie},
  \bibinfo{person}{Alex~X. Lee}, {and} \bibinfo{person}{Sergey Levine}.}
  \bibinfo{year}{2018}\natexlab{}.
\newblock \bibinfo{title}{Visual Foresight: Model-Based Deep Reinforcement
  Learning for Vision-Based Robotic Control}.
\newblock
\newblock


\bibitem[\protect\citeauthoryear{Feinberg, Wan, Stoica, Jordan, Gonzalez, and
  Levine}{Feinberg et~al\mbox{.}}{2018}]%
        {Feinberg_2018}
\bibfield{author}{\bibinfo{person}{Vladimir Feinberg}, \bibinfo{person}{Alvin
  Wan}, \bibinfo{person}{Ion Stoica}, \bibinfo{person}{Michael~I. Jordan},
  \bibinfo{person}{Joseph~E. Gonzalez}, {and} \bibinfo{person}{Sergey Levine}.}
  \bibinfo{year}{2018}\natexlab{}.
\newblock \bibinfo{title}{Model-Based Value Estimation for Efficient Model-Free
  Reinforcement Learning}.
\newblock
\newblock
\showeprint[arxiv]{1803.00101}
\urldef\tempurl%
\url{http://arxiv.org/abs/1803.00101}
\showURL{%
\tempurl}


\bibitem[\protect\citeauthoryear{Freitas, De~Souza, and Almeida}{Freitas
  et~al\mbox{.}}{2009}]%
        {Freitas_2009}
\bibfield{author}{\bibinfo{person}{Fabio Freitas}, \bibinfo{person}{Alberto
  De~Souza}, {and} \bibinfo{person}{Ailson Almeida}.}
  \bibinfo{year}{2009}\natexlab{}.
\newblock \showarticletitle{Prediction-based portfolio optimization model using
  neural networks}.
\newblock \bibinfo{journal}{\emph{Neurocomputing}}  \bibinfo{volume}{72}
  (\bibinfo{date}{06} \bibinfo{year}{2009}), \bibinfo{pages}{2155--2170}.
\newblock


\bibitem[\protect\citeauthoryear{Gal, McAllister, and Rasmussen}{Gal
  et~al\mbox{.}}{2016}]%
        {Gal_2016}
\bibfield{author}{\bibinfo{person}{Yain Gal}, \bibinfo{person}{Rowan
  McAllister}, {and} \bibinfo{person}{Carl~E. Rasmussen}.}
  \bibinfo{year}{2016}\natexlab{}.
\newblock \bibinfo{title}{Improving {PILCO} with {Bayesian} Neural Network
  Dynamics Models}.
\newblock
\newblock


\bibitem[\protect\citeauthoryear{Glosten, Jagannathan, and Runkle}{Glosten
  et~al\mbox{.}}{1993}]%
        {Glosten_1993}
\bibfield{author}{\bibinfo{person}{Lawrence~R Glosten}, \bibinfo{person}{Ravi
  Jagannathan}, {and} \bibinfo{person}{David~E Runkle}.}
  \bibinfo{year}{1993}\natexlab{}.
\newblock \showarticletitle{On the Relation between the Expected Value and the
  Volatility of the Nominal Excess Return on Stocks}.
\newblock \bibinfo{journal}{\emph{Journal of Finance}} \bibinfo{volume}{48},
  \bibinfo{number}{5} (\bibinfo{year}{1993}), \bibinfo{pages}{1779--1801}.
\newblock


\bibitem[\protect\citeauthoryear{Haarnoja, Zhou, Abbeel, and Levine}{Haarnoja
  et~al\mbox{.}}{2018}]%
        {Haarnoja_2018}
\bibfield{author}{\bibinfo{person}{Tuomas Haarnoja}, \bibinfo{person}{Aurick
  Zhou}, \bibinfo{person}{Pieter Abbeel}, {and} \bibinfo{person}{Sergey
  Levine}.} \bibinfo{year}{2018}\natexlab{}.
\newblock \bibinfo{title}{Soft Actor-Critic: Off-Policy Maximum Entropy Deep
  Reinforcement Learning with a Stochastic Actor}.
\newblock
\newblock


\bibitem[\protect\citeauthoryear{Heaton, Polson, and Witte}{Heaton
  et~al\mbox{.}}{2017}]%
        {Heaton_2017}
\bibfield{author}{\bibinfo{person}{J.~B. Heaton}, \bibinfo{person}{N.~G.
  Polson}, {and} \bibinfo{person}{J.~H. Witte}.}
  \bibinfo{year}{2017}\natexlab{}.
\newblock \showarticletitle{Deep learning for finance: deep portfolios}.
\newblock \bibinfo{journal}{\emph{Applied Stochastic Models in Business and
  Industry}} \bibinfo{volume}{33}, \bibinfo{number}{1} (\bibinfo{year}{2017}),
  \bibinfo{pages}{3--12}.
\newblock


\bibitem[\protect\citeauthoryear{Hocquard, Ng, and Papageorgiou}{Hocquard
  et~al\mbox{.}}{2013}]%
        {Hocquard_2013}
\bibfield{author}{\bibinfo{person}{Alexandre Hocquard}, \bibinfo{person}{S.
  Ng}, {and} \bibinfo{person}{N. Papageorgiou}.}
  \bibinfo{year}{2013}\natexlab{}.
\newblock \showarticletitle{A Constant-Volatility Framework for Managing Tail
  Risk}.
\newblock \bibinfo{journal}{\emph{The Journal of Portfolio Management}}
  \bibinfo{volume}{39} (\bibinfo{year}{2013}), \bibinfo{pages}{28--40}.
\newblock


\bibitem[\protect\citeauthoryear{Janner, Fu, Zhang, and Levine}{Janner
  et~al\mbox{.}}{2019}]%
        {Janner_2019}
\bibfield{author}{\bibinfo{person}{Michael Janner}, \bibinfo{person}{Justin
  Fu}, \bibinfo{person}{Marvin Zhang}, {and} \bibinfo{person}{Sergey Levine}.}
  \bibinfo{year}{2019}\natexlab{}.
\newblock \bibinfo{title}{When to Trust Your Model: Model-Based Policy
  Optimization}.
\newblock
\newblock
\showeprint[arxiv]{1906.08253}
\urldef\tempurl%
\url{http://arxiv.org/abs/1906.08253}
\showURL{%
\tempurl}


\bibitem[\protect\citeauthoryear{{Jiang} and {Liang}}{{Jiang} and
  {Liang}}{2016}]%
        {Jiang_2016}
\bibfield{author}{\bibinfo{person}{Zhengyao {Jiang}} {and}
  \bibinfo{person}{Jinjun {Liang}}.} \bibinfo{year}{2016}\natexlab{}.
\newblock \bibinfo{title}{{Cryptocurrency Portfolio Management with Deep
  Reinforcement Learning}}.
\newblock
\newblock
\showeprint{1612.01277}


\bibitem[\protect\citeauthoryear{Kaelbling, Littman, and Moore}{Kaelbling
  et~al\mbox{.}}{1996}]%
        {Kaelbling_1996}
\bibfield{author}{\bibinfo{person}{Leslie~Pack Kaelbling},
  \bibinfo{person}{Michael~L. Littman}, {and} \bibinfo{person}{Andrew~P.
  Moore}.} \bibinfo{year}{1996}\natexlab{}.
\newblock \showarticletitle{Reinforcement Learning: A Survey}.
\newblock \bibinfo{journal}{\emph{Journal of Artificial Intelligence Research}}
   \bibinfo{volume}{4} (\bibinfo{year}{1996}), \bibinfo{pages}{237--285}.
\newblock


\bibitem[\protect\citeauthoryear{Kaiser, Babaeizadeh, Milos, Osinski, Campbell,
  Czechowski, Erhan, Finn, Kozakowsi, Levine, Sepassi, Tucker, and
  Michalewski}{Kaiser et~al\mbox{.}}{2019}]%
        {Kaiser_2019}
\bibfield{author}{\bibinfo{person}{Lukasz Kaiser}, \bibinfo{person}{Mohammad
  Babaeizadeh}, \bibinfo{person}{Piotr Milos}, \bibinfo{person}{Blazej
  Osinski}, \bibinfo{person}{Roy~H. Campbell}, \bibinfo{person}{Konrad
  Czechowski}, \bibinfo{person}{Dumitru Erhan}, \bibinfo{person}{Chelsea Finn},
  \bibinfo{person}{Piotr Kozakowsi}, \bibinfo{person}{Sergey Levine},
  \bibinfo{person}{Ryan Sepassi}, \bibinfo{person}{George Tucker}, {and}
  \bibinfo{person}{Henryk Michalewski}.} \bibinfo{year}{2019}\natexlab{}.
\newblock \bibinfo{title}{Model-Based Reinforcement Learning for {Atari}}.
\newblock
\newblock


\bibitem[\protect\citeauthoryear{Kingma and Ba}{Kingma and Ba}{2014}]%
        {kingma2014method}
\bibfield{author}{\bibinfo{person}{Diederik Kingma} {and}
  \bibinfo{person}{Jimmy Ba}.} \bibinfo{year}{2014}\natexlab{}.
\newblock \bibinfo{title}{Adam: A Method for Stochastic Optimization}.
\newblock
\newblock


\bibitem[\protect\citeauthoryear{{Kumar}, {Todorov}, and {Levine}}{{Kumar}
  et~al\mbox{.}}{2016}]%
        {Kumar_2016}
\bibfield{author}{\bibinfo{person}{Vikash {Kumar}}, \bibinfo{person}{Emanuel
  {Todorov}}, {and} \bibinfo{person}{Sergey {Levine}}.}
  \bibinfo{year}{2016}\natexlab{}.
\newblock \bibinfo{title}{Optimal control with learned local models:
  Application to dexterous manipulation}.
\newblock , \bibinfo{numpages}{378--383}~pages.
\newblock


\bibitem[\protect\citeauthoryear{Levine and Koltun}{Levine and Koltun}{2013}]%
        {Levine_Koltun_2013}
\bibfield{author}{\bibinfo{person}{Sergey Levine} {and}
  \bibinfo{person}{Vladlen Koltun}.} \bibinfo{year}{2013}\natexlab{}.
\newblock \bibinfo{title}{Guided Policy Search}.
\newblock , \bibinfo{numpages}{9}~pages.
\newblock
\urldef\tempurl%
\url{http://proceedings.mlr.press/v28/levine13.html}
\showURL{%
\tempurl}


\bibitem[\protect\citeauthoryear{Li, Li, Zhan, and Liu}{Li
  et~al\mbox{.}}{2019}]%
        {Li_2019}
\bibfield{author}{\bibinfo{person}{Xinyi Li}, \bibinfo{person}{Yinchuan Li},
  \bibinfo{person}{Yuancheng Zhan}, {and} \bibinfo{person}{Xiao-Yang Liu}.}
  \bibinfo{year}{2019}\natexlab{}.
\newblock \bibinfo{title}{Optimistic Bull or Pessimistic Bear: Adaptive Deep
  Reinforcement Learning for Stock Portfolio Allocation}.
\newblock
\newblock


\bibitem[\protect\citeauthoryear{{Liang et al.}}{{Liang et al.}}{2018}]%
        {Liang_2018}
\bibfield{author}{\bibinfo{person}{{Liang et al.}}}
  \bibinfo{year}{2018}\natexlab{}.
\newblock \bibinfo{title}{Adversarial Deep Reinforcement Learning in Portfolio
  Management}.
\newblock
\newblock
\showeprint[arxiv]{1808.09940}~[q-fin.PM]


\bibitem[\protect\citeauthoryear{Lillicrap, Hunt, Pritzel, Heess, Erez, Tassa,
  Silver, and Wierstra}{Lillicrap et~al\mbox{.}}{2016}]%
        {Lillicrap_2016}
\bibfield{author}{\bibinfo{person}{Timothy~P. Lillicrap},
  \bibinfo{person}{Jonathan~J. Hunt}, \bibinfo{person}{Alexander Pritzel},
  \bibinfo{person}{Nicolas Heess}, \bibinfo{person}{Tom Erez},
  \bibinfo{person}{Yuval Tassa}, \bibinfo{person}{David Silver}, {and}
  \bibinfo{person}{Daan Wierstra}.} \bibinfo{year}{2016}\natexlab{}.
\newblock \bibinfo{title}{Continuous control with deep reinforcement learning}.
\newblock
\newblock


\bibitem[\protect\citeauthoryear{Liu, Liu, Zhao, Pan, and Liu}{Liu
  et~al\mbox{.}}{2020}]%
        {Liu_2020}
\bibfield{author}{\bibinfo{person}{Yang Liu}, \bibinfo{person}{Qi Liu},
  \bibinfo{person}{Hongke Zhao}, \bibinfo{person}{Zhen Pan}, {and}
  \bibinfo{person}{Chuanren Liu}.} \bibinfo{year}{2020}\natexlab{}.
\newblock \bibinfo{title}{Adaptive Quantitative Trading: an Imitative Deep
  Reinforcement Learning Approach}.
\newblock
\newblock


\bibitem[\protect\citeauthoryear{Markowitz}{Markowitz}{1952}]%
        {Markowitz_1952}
\bibfield{author}{\bibinfo{person}{Harry Markowitz}.}
  \bibinfo{year}{1952}\natexlab{}.
\newblock \showarticletitle{Portfolio Selection}.
\newblock \bibinfo{journal}{\emph{The Journal of Finance}} \bibinfo{volume}{7},
  \bibinfo{number}{1} (\bibinfo{date}{March} \bibinfo{year}{1952}),
  \bibinfo{pages}{77--91}.
\newblock


\bibitem[\protect\citeauthoryear{Mnih, Kavukcuoglu, Silver, Rusu, Veness,
  Bellemare, Graves, Riedmiller, Fidjeland, Ostrovski, Petersen, Beattie,
  Sadik, Antonoglou, King, Kumaran, Wierstra, Legg, and Hassabis}{Mnih
  et~al\mbox{.}}{2015}]%
        {Mnih_2015}
\bibfield{author}{\bibinfo{person}{Volodymyr Mnih}, \bibinfo{person}{Koray
  Kavukcuoglu}, \bibinfo{person}{David Silver}, \bibinfo{person}{Andrei Rusu},
  \bibinfo{person}{Joel Veness}, \bibinfo{person}{Marc Bellemare},
  \bibinfo{person}{Alex Graves}, \bibinfo{person}{Martin Riedmiller},
  \bibinfo{person}{Andreas Fidjeland}, \bibinfo{person}{Georg Ostrovski},
  \bibinfo{person}{Stig Petersen}, \bibinfo{person}{Charles Beattie},
  \bibinfo{person}{Amir Sadik}, \bibinfo{person}{Ioannis Antonoglou},
  \bibinfo{person}{Helen King}, \bibinfo{person}{Dharshan Kumaran},
  \bibinfo{person}{Daan Wierstra}, \bibinfo{person}{Shane Legg}, {and}
  \bibinfo{person}{Demis Hassabis}.} \bibinfo{year}{2015}\natexlab{}.
\newblock \showarticletitle{Human-level control through deep reinforcement
  learning}.
\newblock \bibinfo{journal}{\emph{Nature}}  \bibinfo{volume}{518}
  (\bibinfo{date}{02} \bibinfo{year}{2015}), \bibinfo{pages}{529--33}.
\newblock


\bibitem[\protect\citeauthoryear{Moerland, Broekens, and Jonker}{Moerland
  et~al\mbox{.}}{2020}]%
        {moerland2020modelbased}
\bibfield{author}{\bibinfo{person}{Thomas~M. Moerland}, \bibinfo{person}{Joost
  Broekens}, {and} \bibinfo{person}{Catholijn~M. Jonker}.}
  \bibinfo{year}{2020}\natexlab{}.
\newblock \bibinfo{title}{Model-based Reinforcement Learning: A Survey}.
\newblock
\newblock
\showeprint[arxiv]{2006.16712}~[cs.LG]


\bibitem[\protect\citeauthoryear{Nagabandi, Kahn, Fearing, and
  Levine}{Nagabandi et~al\mbox{.}}{2018}]%
        {Nagabandi_2018}
\bibfield{author}{\bibinfo{person}{Anusha Nagabandi}, \bibinfo{person}{Gregory
  Kahn}, \bibinfo{person}{Ronald Fearing}, {and} \bibinfo{person}{Sergey
  Levine}.} \bibinfo{year}{2018}\natexlab{}.
\newblock \bibinfo{title}{Neural Network Dynamics for Model-Based Deep
  Reinforcement Learning with Model-Free Fine-Tuning}.
\newblock , \bibinfo{numpages}{7559--7566}~pages.
\newblock
\urldef\tempurl%
\url{https://doi.org/10.1109/ICRA.2018.8463189}
\showDOI{\tempurl}


\bibitem[\protect\citeauthoryear{Niaki and Hoseinzade}{Niaki and
  Hoseinzade}{2013}]%
        {Niaki2013}
\bibfield{author}{\bibinfo{person}{Seyed Niaki} {and} \bibinfo{person}{Saeid
  Hoseinzade}.} \bibinfo{year}{2013}\natexlab{}.
\newblock \showarticletitle{Forecasting S\&P 500 index using artificial neural
  networks and design of experiments}.
\newblock \bibinfo{journal}{\emph{Journal of Industrial Engineering
  International}}  \bibinfo{volume}{9} (\bibinfo{date}{02}
  \bibinfo{year}{2013}).
\newblock


\bibitem[\protect\citeauthoryear{Noureldin and Shephard}{Noureldin and
  Shephard}{2012}]%
        {Noureldin12multivariatehigh-frequency-based}
\bibfield{author}{\bibinfo{person}{Diaa Noureldin} {and} \bibinfo{person}{Neil
  Shephard}.} \bibinfo{year}{2012}\natexlab{}.
\newblock \bibinfo{title}{Multivariate high-frequency-based volatility (HEAVY)
  models}.
\newblock , \bibinfo{numpages}{907--933}~pages.
\newblock


\bibitem[\protect\citeauthoryear{Oh, Guo, Lee, Lewis, and Singh}{Oh
  et~al\mbox{.}}{2015}]%
        {Oh_2015}
\bibfield{author}{\bibinfo{person}{Junhyuk Oh}, \bibinfo{person}{Xiaoxiao Guo},
  \bibinfo{person}{Honglak Lee}, \bibinfo{person}{Richard Lewis}, {and}
  \bibinfo{person}{Satinder Singh}.} \bibinfo{year}{2015}\natexlab{}.
\newblock \bibinfo{title}{Action-Conditional Video Prediction using Deep
  Networks in {Atari} Games}.
\newblock
\newblock


\bibitem[\protect\citeauthoryear{Perchet, Leote~de Carvalho, Heckel, and
  Moulin}{Perchet et~al\mbox{.}}{2016}]%
        {Perchet_2016}
\bibfield{author}{\bibinfo{person}{Romain Perchet}, \bibinfo{person}{Raul
  Leote~de Carvalho}, \bibinfo{person}{Thomas Heckel}, {and}
  \bibinfo{person}{Pierre Moulin}.} \bibinfo{year}{2016}\natexlab{}.
\newblock \bibinfo{title}{Predicting the Success of Volatility Targeting
  Strategies: Application to Equities and Other Asset Classes}.
\newblock , \bibinfo{numpages}{21--38}~pages.
\newblock


\bibitem[\protect\citeauthoryear{Pong, Gu, Dalal, and Levine}{Pong
  et~al\mbox{.}}{2018}]%
        {Pong_2018}
\bibfield{author}{\bibinfo{person}{Vitchyr Pong}, \bibinfo{person}{Shixiang
  Gu}, \bibinfo{person}{Murtaza Dalal}, {and} \bibinfo{person}{Sergey Levine}.}
  \bibinfo{year}{2018}\natexlab{}.
\newblock \bibinfo{title}{Temporal Difference Models: Model-Free Deep {RL} for
  Model-Based Control}.
\newblock
\newblock
\showeprint[arxiv]{1802.09081}
\urldef\tempurl%
\url{http://arxiv.org/abs/1802.09081}
\showURL{%
\tempurl}


\bibitem[\protect\citeauthoryear{Salhi, Deaconu, Lejay, Champagnat, and
  Navet}{Salhi et~al\mbox{.}}{2015}]%
        {Salhi_2016}
\bibfield{author}{\bibinfo{person}{Khaled Salhi}, \bibinfo{person}{Madalina
  Deaconu}, \bibinfo{person}{Antoine Lejay}, \bibinfo{person}{Nicolas
  Champagnat}, {and} \bibinfo{person}{Nicolas Navet}.}
  \bibinfo{year}{2015}\natexlab{}.
\newblock \bibinfo{title}{Regime switching model for financial data: empirical
  risk analysis}.
\newblock
\newblock


\bibitem[\protect\citeauthoryear{Sutton and Barto}{Sutton and Barto}{2018}]%
        {SuttonBarto_2018}
\bibfield{author}{\bibinfo{person}{Richard~S. Sutton} {and}
  \bibinfo{person}{Andrew~G. Barto}.} \bibinfo{year}{2018}\natexlab{}.
\newblock \bibinfo{booktitle}{\emph{Reinforcement Learning: An Introduction}}.
\newblock \bibinfo{publisher}{The MIT Press}, \bibinfo{address}{The MIT Press}.
\newblock


\bibitem[\protect\citeauthoryear{van Hasselt, Hessel, and Aslanides}{van
  Hasselt et~al\mbox{.}}{2019}]%
        {van_Hasselt_2018}
\bibfield{author}{\bibinfo{person}{Hado van Hasselt}, \bibinfo{person}{Matteo
  Hessel}, {and} \bibinfo{person}{John Aslanides}.}
  \bibinfo{year}{2019}\natexlab{}.
\newblock \bibinfo{title}{When to use parametric models in reinforcement
  learning?}
\newblock
\newblock
\showeprint[arxiv]{1906.05243}
\urldef\tempurl%
\url{http://arxiv.org/abs/1906.05243}
\showURL{%
\tempurl}


\bibitem[\protect\citeauthoryear{Wang and Zhou}{Wang and Zhou}{2019}]%
        {Wang_2019}
\bibfield{author}{\bibinfo{person}{Haoran Wang} {and} \bibinfo{person}{Xun~Yu
  Zhou}.} \bibinfo{year}{2019}\natexlab{}.
\newblock \bibinfo{title}{Continuous-Time Mean-Variance Portfolio Selection: A
  Reinforcement Learning Framework}.
\newblock
\newblock
\showeprint[arxiv]{1904.11392}~[q-fin.PM]


\bibitem[\protect\citeauthoryear{Xiong, Liu, Zhong, Yang, and Walid}{Xiong
  et~al\mbox{.}}{2019}]%
        {Xiong_2018}
\bibfield{author}{\bibinfo{person}{Zhuoran Xiong}, \bibinfo{person}{Xiao-Yang
  Liu}, \bibinfo{person}{Shan Zhong}, \bibinfo{person}{Hongyang Yang}, {and}
  \bibinfo{person}{Anwar Walid}.} \bibinfo{year}{2019}\natexlab{}.
\newblock \bibinfo{title}{Practical Deep Reinforcement Learning Approach for
  Stock Trading}.
\newblock
\newblock


\bibitem[\protect\citeauthoryear{Ye, Pei, Wang, Chen, Zhu, Xiao, and Li}{Ye
  et~al\mbox{.}}{2020}]%
        {Ye_2020}
\bibfield{author}{\bibinfo{person}{Yunan Ye}, \bibinfo{person}{Hengzhi Pei},
  \bibinfo{person}{Boxin Wang}, \bibinfo{person}{Pin-Yu Chen},
  \bibinfo{person}{Yada Zhu}, \bibinfo{person}{Ju Xiao}, {and}
  \bibinfo{person}{Bo Li}.} \bibinfo{year}{2020}\natexlab{}.
\newblock \showarticletitle{Reinforcement-Learning Based Portfolio Management
  with Augmented Asset Movement Prediction States}. In
  \bibinfo{booktitle}{\emph{AAAI}}. \bibinfo{publisher}{AAAI},
  \bibinfo{address}{New York}, \bibinfo{pages}{1112--1119}.
\newblock
\urldef\tempurl%
\url{https://aaai.org/ojs/index.php/AAAI/article/view/5462}
\showURL{%
\tempurl}


\bibitem[\protect\citeauthoryear{Yu, Lee, Kulyatin, Shi, and Dasgupta}{Yu
  et~al\mbox{.}}{2019}]%
        {Yu_2019}
\bibfield{author}{\bibinfo{person}{Pengqian Yu}, \bibinfo{person}{Joon~Sern
  Lee}, \bibinfo{person}{Ilya Kulyatin}, \bibinfo{person}{Zekun Shi}, {and}
  \bibinfo{person}{Sakyasingha Dasgupta}.} \bibinfo{year}{2019}\natexlab{}.
\newblock \bibinfo{title}{Model-based Deep Reinforcement Learning for Financial
  Portfolio Optimization}.
\newblock
\newblock


\bibitem[\protect\citeauthoryear{Zheng, Li, and Xu}{Zheng
  et~al\mbox{.}}{2019}]%
        {Zheng_2019}
\bibfield{author}{\bibinfo{person}{Kai Zheng}, \bibinfo{person}{Yuying Li},
  {and} \bibinfo{person}{Weidong Xu}.} \bibinfo{year}{2019}\natexlab{}.
\newblock \bibinfo{title}{Regime switching model estimation: spectral
  clustering hidden Markov model}.
\newblock
\newblock


\bibitem[\protect\citeauthoryear{{Zhengyao et al.}}{{Zhengyao et al.}}{2017}]%
        {Zhengyao_2017}
\bibfield{author}{\bibinfo{person}{{Zhengyao et al.}}}
  \bibinfo{year}{2017}\natexlab{}.
\newblock \bibinfo{title}{Reinforcement Learning Framework for the Financial
  Portfolio Management Problem}.
\newblock
\newblock
\showeprint{1706.10059v2}


\end{thebibliography}

\end{document}